%% file: neurips_2020.tex
\documentclass{article}

\usepackage[preprint]{neurips_2020}

\usepackage[utf8]{inputenc} 
\usepackage[T1]{fontenc}    
\usepackage{amsfonts}       
\usepackage{booktabs}       
\usepackage{hyperref}       
\usepackage{url}            
\usepackage{microtype}      
\usepackage{nicefrac}       
\usepackage{paralist}       
\usepackage{siunitx}        

\usepackage{amssymb, amsfonts, amsmath, mathtools, blkarray, tikz-cd, float, amsthm, caption, algorithmicx, algorithm, wrapfig}
\usepackage[noend]{algpseudocode}

\newtheorem{remark}{Remark}

\newtheorem*{theorem*}{Theorem}
\DeclareMathOperator*{\argmax}{arg\,max}
\DeclareMathOperator*{\argmin}{arg\,min}
\newcommand{\floor}[1]{\lfloor #1 \rfloor}

\newcommand{\norm}[1]{\left\lVert#1\right\rVert}

\newcommand*\Let[2]{\State #1 $\gets$ #2}
\algrenewcommand\algorithmicrequire{\textbf{Precondition:}}
\algrenewcommand\algorithmicensure{\textbf{Postcondition:}}

\title{Geometric feature performance under downsampling for EEG classification tasks}

\author{%
  Bryan Bischof \\
  Stitch Fix Algorithms\\
  Berkeley, CA \\
  \texttt{bryan.bischof@gmail.com} \\
  \And
  Eric Bunch \\
  American Family Insurance \\
  Madison, WI  \\
  \texttt{eric.a.bunch@gmail.com} \\
}

\begin{document}

\maketitle

\begin{abstract}
  We experimentally investigate a collection of feature engineering pipelines for use with a CNN for classifying eyes-open or eyes-closed from electroencephalogram (EEG) time-series from the Bonn dataset. Using the Takens' embedding--a geometric representation of time-series--we construct simplicial complexes from EEG data. We then compare $\epsilon$-series of Betti-numbers and $\epsilon$-series of graph spectra (a novel construction)--two topological invariants of the latent geometry from these complexes--to raw time series of the EEG to fill in a gap in the literature for benchmarking. These methods, inspired by Topological Data Analysis, are used for feature engineering to capture local geometry of the time-series. Additionally, we test these feature pipelines' robustness to downsampling and data reduction. This paper seeks to establish clearer expectations for both time-series classification via geometric features, and how CNNs for time-series respond to data of degraded resolution.
\end{abstract}

\input{final_section_files/introduction}

\input{final_section_files/methods}

\input{final_section_files/experiments}

\input{final_section_files/outcomes}

\bibliographystyle{plainnat}
\bibliography{neurips_2020}

\input{final_section_files/acknowledgements_and_impacts}

\appendix
\input{final_section_files/appendices}

\end{document}

%% file: final_section_files/introduction.tex
	\section*{Introduction} Topological Data Analysis (TDA) (\cite{computingPH},\cite{topo_persis}) has gained much attention due to applications for data analysis and machine learning. In particular, persistent homology (\cite{Scopigno04persistencebarcodes}, \cite{Edelsbrunner2000TopologicalPA}) has been leveraged for machine learning purposes in numerous tasks. The methods attempt to describe the shape of the data in a latent space particularly amenable to feature engineering. The efficacy of topological features has been demonstrated in various tasks (\cite{Chazal2017AnIT}, \cite{TDAforArrhythmia}). 

In this paper, we investigate the performance of various topological feature engineering approaches for EEG time-series classification using one dimensional CNN as the classifiers. While CNN architectures are heavily experimented on, less research has explored models for feature engineering using modern geometric techniques (\cite{gcrn, geom_dl}). Usually, CNNs are trained on the raw time-series data where a convolutional kernels of a fixed sizes and strides are applied to the series with moving windows to compute higher-order features. Persistent homology of the {\em Takens' embedding} provides one geometric procedure to engineer features for a time-series (\cite{Umeda2017}). 

For a time-series, the $k$-dimensional {\em Takens' embedding} of the time-series is the Euclidean embedding of points defined by a sliding window of size $k$---this provides a point-cloud representation of the time-series. From this point cloud the common TDA approach to compute homology of the $\epsilon$-Rips complex provides persistent features. Both the raw series and the persistent features can be exploited for machine learning tasks. As a first step towards better understanding of feature engineering on time-series, we compare the performance of these two approaches for the classification task.

Furthermore, we propose a novel geometric method beyond homology theories utilizing eigenvalues of sequences of graph Laplacians. Again, we utilize the point-cloud representation's $\epsilon$-neighbor graph, and compute the normalized graph Laplacians' eigenvalues. Density counts of these eigenvalues are encoded as $m$ discrete $\epsilon$-series. We demonstrate the superiority of our approach over the homological features and compare to the raw time-series via classification experiments while keeping the classifier architecture fixed.   
	
Generally, neural networks are optimized via network features like batch size, learning rate, kernels, pooling layers, and the like. Some papers have experimented with data resizing to improve training time --- in (\cite{DAWNBench}) they introduced dynamic resizing with progressive resolution; (\cite{DBLP}) have investigated the CNN robustness under noise. We have not found in the literature examinations of explicit downsampling algorithms' effects. To this effect, we study the performance of the feature engineering methods across multiple regimes. We vary features, time-series resolution, effective resolutions, and compare degradation across the feature types and downsampling methods–––a practice common in the signal processing literature but less so in classification of time-series. Note that the encoding methods in this paper need not only apply to sequential collections.

	The principal contributions of this work include:
	\begin{itemize}
		\item Introduce a new feature engineering technique utilizing latent geometric properties of the time series.
		\item Apply the theory and methods of downsampling to time-series classification problem. 
		\item Propose and demonstrate a comparison framework and baseline results for time series clustering via varying features and CNN architectures. 
	\end{itemize}
	
	\subsection*{Comments and Caveats}

	In machine learning tasks, especially those with more complicated models, it is essential to attempt to establish a baseline, a set of target metrics, and a comparison pipeline for apples-to-apples evaluation\textemdash in some literature this is called boat-racing, or horse-racing. During our literature review of TDA methods for the EEG prediction dataset, we did not find sufficient comparators and thus integrate this into our experimentation goals, whence the large volume of experiments considered in this paper.

	We do not treat the neuroscience topics necessary for a deeper investigation of EEG technology or seizures. We treat the dataset as a mathematical task, and focus instead on the models and methods in machine learning. In particular, as highlighted in (\cite{GCN-EEG}) EEG datasets are structural time series, i.e. physical geometry of data collection correlates with relationships between series. We use only local readings thus reducing this covariance structure. Furthermore, train and test splitting can be challenging as several samples are taken from the same patient. For these, and other reasons, we make no comparison to SOTA networks. \textbf{The methods and architectures covered in this work are not state-of-the-art in terms of performance.}

	\subsection*{This work in context}

	Generally, CNN architectures are optimized via network features like batch size, learning rate, kernels, pooling layers, and the like. Some papers have experimented with resizing to improve training time --- in (\cite{DAWNBench}) they introduced dynamic resizing with progressive resolution. Others (\cite{DBLP}) have investigated the robustness of CNN performance under image degradation due to noise, but we've not found in the literature examinations of explicit downsampling algorithms' effects. There is renewed interest in small models over the last few years in other ML tasks; for time-series prediction this paper provides one such framework for pursuing those ends.

	For a collection classification task, the encoding of the data itself is also at opportunity for tuning. While common approaches to sequential data sets use RNNs and LSTMs to take advantage of data characteristics like autoregressive features, the encoding methods in this paper need not only apply to sequential collections. 

	We wish to highlight that while time series classification is well studied and has clear and effective baselines (\cite{TSclass}), EEG classifiers are of great value and remain elusive.

%% file: final_section_files/methods.tex
\section{Methods}

We use two geometric methods for feature engineering, persistent Betti numbers and persistent Spectra. The first is a development of \cite{Umeda2017}, the second is a new construction.

\textbf{Takens' Embedding}\label{sec:takens} Given a time series, $\{t_i \}_{i=0}^n \subset \mathbb{R}$, and a function $f: \mathbb{R} \rightarrow \mathbb{R}$. Then the \emph{Takens' embedding} with window size $m$, denoted $T^m$, is the collection of points in $\mathbb{R}^m$ given by
	\begin{equation*}
	\left\{  
		\left[ f(t_0), f(t_1), \dots, f(t_{m-1})  \right], 
		\left[ f(t_1), f(t_2), \dots, f(t_{m}) \right],  
		\dots,  
		\left[ f(t_{n-m+1}), f(t_{n - m +1}), \dots f(t_{n}) \right] 
	\right\} 
	\end{equation*}
\noindent That is, $T^m$ is the collection of points in $\mathbb{R}^m$ given by taking sliding windows over the time series $f(t_i)$. For volatile series, smooth transformations on this geometry are known to preserve properties of the time series, such as the dimension of chaotic attractor, and Lyapunov exponents of the dynamics.

\textbf{$\epsilon$-neighbor graph} Given a finite subset $X \subset \mathbb{R}^m$ with $n$ points, and a real number $\epsilon \geq 0$ we form the $\epsilon$-graph $G_{\epsilon}(X)$ with nodes of $G_{\epsilon}(X)$ indexed by the elements of $X$, and edges $v_x\rightarrow v_y$ for $x, y \in X$ if and only if $\norm{x - y} < \epsilon$.
 
\textbf{Persistent Homology and Time Series Analysis}\label{sec:ph_ts} In (\cite{Umeda2017}), the author calculates persistent Betti numbers(c.f. the appendix: \ref{ph} for definitions) of EEG time series signals via the aforementioned Persistent Betti-number model pipeline before feeding the output into a CNN for an eyes open/closed prediction task. In particular, for time series $\{ f(t_i)  \}_{i=0}^n$, consider $T^m$ its Takens' embedding as the enumerated subset in $\mathbb{R}^m$, then compute $\beta_{k}(\epsilon)$ for all $0 \leq k < m$ and $\epsilon \in [0,r]$.

\textbf{Graph Laplacians} Spectral graph theory is an integral facet of graph theory (\cite{chung1997spectral}) and one of the key objects of this theory is the Laplacian matrix of a graph, as well as its eigenvalues. 
We assume all graphs are undirected and simple. For a graph $G$, let $A$ and $D$ be the adjacency matrix and the degree matrix of $G$ respectively.

The \emph{Laplacian} of $G$ is defined to be $L = D - A$. The \emph{normalized Laplacian} of $G$ is then defined to be $\tilde{L}  = D^{-1/2} A D^{-1/2}.$

Denote the eigenvalues(or \emph{spectrum}) of $\tilde{L}$ by  $0=\lambda_0 \leq \lambda_1 \leq \cdots \leq \lambda_{n-1}$. You may recall:

[\textbf{Lemma 1.7}, \cite{chung1997spectral}]\label{thm:graphspec}
For a graph $G$ with $n$ vertices, we have that
\begin{enumerate}
	\item $0 \leq \lambda_i \leq 2$, with $\lambda_0 = 0$.  
	\item If $G$ is connected, then $\lambda_1 > 0$. If $\lambda_i = 0$ and $\lambda_{i+1} \neq 0$ then $G$ has exactly $i+1$ connected components.
\end{enumerate}

\textbf{Persistent Laplacian Eigenvalues for Time Series Analysis}\label{sec:laplacians_ts} Denote by $\tilde{L}_{\epsilon}(X)$ the normalized Laplacian of $G_{\epsilon}(X)$. Define $\hat{\lambda}_{\epsilon}(X) = [ \lambda_{\epsilon}(X)_0, \lambda_{\epsilon}(X)_1, \dots \lambda_{\epsilon}(X)_{n-1} ]$ to be the vector of eigenvalues of $\tilde{L}_{\epsilon}(X)$, in ascending order: $0 = \lambda_{\epsilon}(X)_0 \leq \lambda_{\epsilon}(X)_1 \leq \cdots \leq \lambda_{\epsilon}(X)_{n-1} \leq 2$. When the context is understood, we will drop the designation $(X)$ in the above notations; e.g. $G_{\epsilon}$ or $\lambda_{\epsilon 0}$.

Let $I$ be an interval, $\hat{v} = [v_0, v_1, \dots, v_{n-1}]$ be a vector, and define 

\begin{equation}
\textbf{count}_{I}(v) : = \#\{ v_i \mid v_i \in I  \}.
\end{equation}
For a given interval $[0, r]$ (this will be our range of resolutions), and a finite collection of real numbers $0 = \tau_0 < \tau_1 < \cdots < \tau_k = 2 + \delta $ for any fixed $\delta$, such that $0<\delta\in\mathbb{R}$. For $\epsilon \in [0, r]$, define:
\begin{equation}
\mu_j(\epsilon) := 
	\textbf{count}_{[\tau_j, \tau_{j + 1})}(\hat{\lambda_{\epsilon}}) \text{  for } 0 \leq j \leq k - 1 
\end{equation}

\noindent That is, $\mu_j(\epsilon)$ counts the number of eigenvalues of $\tilde{L}_{\epsilon}$ that lie between $\tau_j$ and $\tau_{j + 1}$. Observe that $\textbf{count}_{[0, 0]}(\hat{\lambda_{\epsilon}})$ is equal to the number of connected components of $G_{\epsilon}$. We will view the collection $\{ \mu_j \}$ as a collection  of $j$ real-valued functions with domain $[0, r]$. We refer to the collection of $\mu_j$'s as \emph{persistent Laplacian eigenvalues}. Given a time series $\{ f(t_i)  \}_{i=0}^n$ we form $T^m$, and compute $\{ \mu_j \}_{j=0}^l$ for some choice of $\tau_0, \dots \tau_l$.
\newpage
 \subsection*{Overview of model pipelines}
	
\begin{wrapfigure}{r}{0.5\textwidth}
  \centering 
  \label{fig:pers_lpn_eigs}
    \includegraphics[width=.48\textwidth]{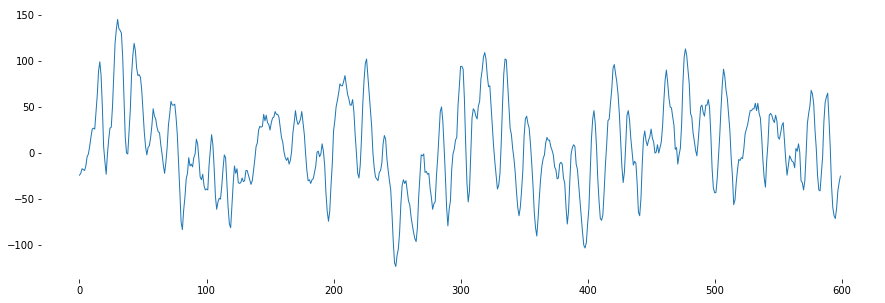}
    \caption{Raw time series from a segment of time where the patient's eyes were closed, a segment of 600 time steps, not downsampled. The x-axis is in units of time-steps and the y-axis is $\mu V$ amplitude}
    \label{fig:raw_time_series}
    
    \includegraphics[width=.48\textwidth]{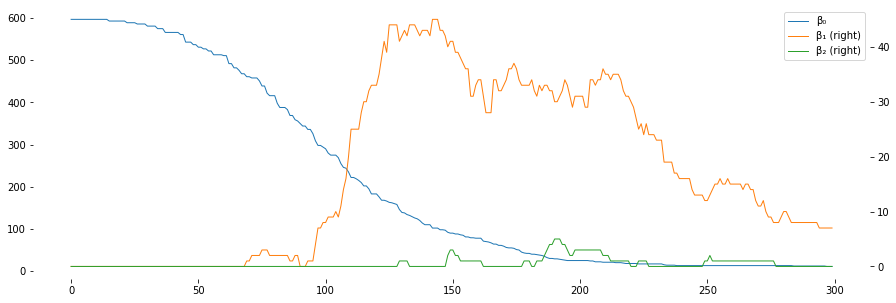}
    \caption{$\beta_j(\epsilon)$ computed for the time series shown in Fig. \ref{fig:raw_time_series}. The x-axis is in units of $\epsilon$-steps and the y-axis is the counts identified by legend.}
    \label{fig:pers_betti_nums}
    
    \includegraphics[width=.48\textwidth]{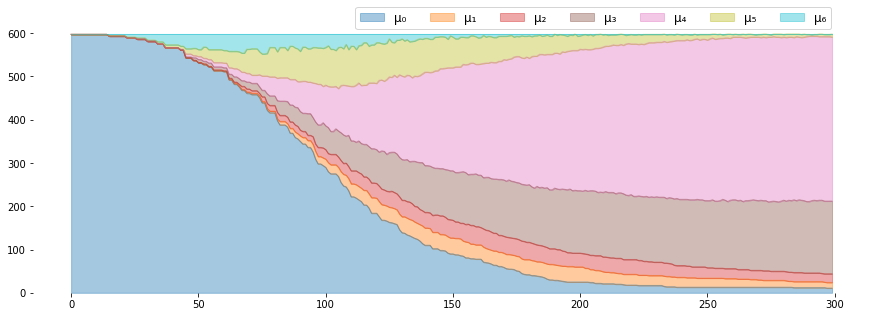}
    \caption{Area plot of $\mu_j(\epsilon)$, i.e. percentage of binned eigenvalues,  for the time series in Fig. \ref{fig:raw_time_series}. The x-axis is in units of $\epsilon$-steps and the y-axis is the counts.}
\end{wrapfigure}

\textbf{Raw Time Series Features:} We feed the sequential time-series values into our CNN architecture.

\textbf{Persistent Betti Numbers Features:}	We encode each time series with the `k-step` Takens' embedding into $\mathbb{R}^k$. This point cloud's $\epsilon$-neighbor graph generates the Vietoris-Rips filtration up to dimension 3. The order of the degree $n$ simplicial homology (or $n$'th Betti number) is computed for each $\epsilon$ neighbor complex, and encoded as $n$ discrete $\epsilon$-series. We feed the sequential $\epsilon$-series values --- each on their own channel --- into our CNN architecture.

\textbf{Persistent Laplacian Eigenvalue Features:}	We encode each time series with the 'k-step' Takens' embedding into $\mathbb{R}^k$. This point cloud's $\epsilon$-neighbor graph is collected and it's normalized graph Laplacians are computed. The eigenvalues of these Laplacians are computed and bucketed into a partition of $m$ buckets. The counts of eigenvalues in each bucket are encoded as $m$ discrete $\epsilon$-series. We feed the sequential $\epsilon$-series values --- each on their own channel --- into our CNN architecture.

\textbf{Experimental design:}
In each of the pipelines, we prepend the model pipeline with a downsampling step using one of three downsampling algorithms and several downsampling resolutions(c.f. the appendix \ref{downsampling}). Our design matrix consists of the three downsampling methods applied to $\left\lbrace 200, 300, 400, 500, 600\right\rbrace$ initial data lengths downsampled by multiples of $50$. Each of these initial data sets are fed through each of the model pipelines and subsequent CNNs. We use cross-validation and accuracy to evaluate the performance.

%% file: final_section_files/experiments.tex
\section{Experiments}

\textbf{Data set and classification task} The data used in this work are time series EEG signals, and are provided by the University of Bonn, explored in \cite{eeg_data}. This data set is comprised of five sets (labeled A-E), each containing 100 single-channel EEG segments 23.6 seconds in duration, with 4097 observations. The segments were hand selected from within a continuous single-channel EEG recording, chosen for absence of artifacts as well as fulfilling a stationarity condition. Set E contains segments of seizure activity, sets C and D are taken from epilepsy patients during an interval where no seizure activity was occurring, and sets A and B are observations from non epilepsy diagnosed patients. The observations in set A occur during times when the patient's eyes were open, while those in set B occur during times when the patient's eyes were closed. We study the classification task of A vs. B., i.e. eyes open or closed.

\textbf{CNN architectures} All of the prediction algorithms  used in this paper are CNNs, each with two sets of one dimensional convolution and max pool layers, followed by a fully connected layer to predict the class label. Architecture parameters are vectors representing $\left<\textrm{input},\textrm{channels},\textrm{factor},\textrm{kernel1 size},\textrm{kernel2 size}\right>$; $\left<\textrm{res},1,5,(\textrm{res}/600)*18,2\right>,\left<300,3,7,6,2\right>,\left<300,7,3,6,2\right>$ for raw time-series, Betti numbers, and eigenvalues respectively. \texttt{factor} refers to the multiplicative factor of input channels to output channels in each convolution layer, and \texttt{res} refers to the resolution to which the time series was downsampled to. Stride and dilation in both conv layers are 1 for each model; first pooling layer has size 7, second has size 3; each model is trained for 10 epochs. Longer training was explored but in testing, more than ten epochs showed little improvement in performance.

\begin{figure}[ht]
\centering
\includegraphics[width=0.75\textwidth]{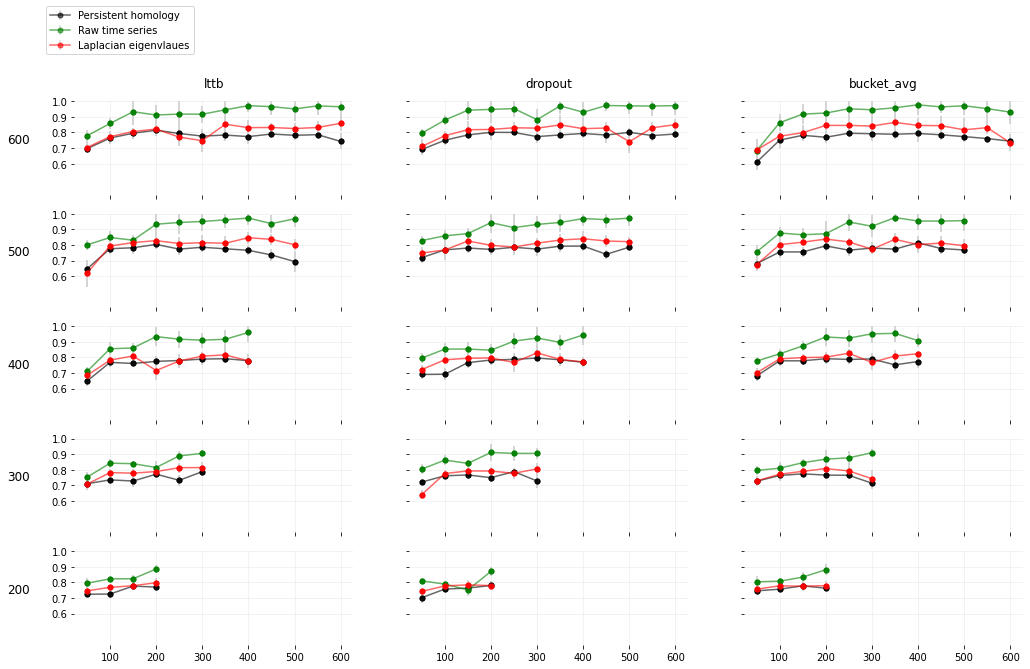}
\caption{Experiment results for non dynamic bucketing downsampling methods. Diagram rows correspond to initial length of time-series segment, with each point reflecting number of points after downsampling. Within rows, y-axis is accuracy on binary cross-entropy, x-axis is the number of points in the time-series samples.}
\label{fig:perf_non_dynamic}
\end{figure}

\textbf{Experiments and Results} For a given observation length, downsampling method, and downsampling rate, we run 10-fold cross validation for each of the three prediction methods described. The average and standard deviation of accuracy is recorded and displayed in Figure \ref{fig:perf_non_dynamic} for non-dynamic bucketing downsampling methods, and Figure \ref{fig:perf_dynamic} for dynamic bucketing downsampling methods.

%% file: final_section_files/outcomes.tex
\section{Outcomes}
		We've explored a collection of `experiments' in training and testing CNNs built on EEG data to predict if a patient's eyes are open or closed. The aforementioned experiments primarily sought to establish performance comparisons while varying the feature engineering choice, the chunk size, and the downsampling resolution used.

		We established a baseline performance using a raw time series feature set and reproduced performance in (\cite{Umeda2017}) to compare to this baseline. We saw that the performance of this baseline actually outperforms the TDA feature engineered experiment as reported. This suggests that for a task of this type the TDA approach is not SOTA, but may hold value in other regimes, or under more specific hyperparameter tuning. Building on these results, we explored the novel geometric feature engineering method of persistent eigenvalues of the Laplacian. This method also outperforms TDA, but does not outperform the raw time series experiment.

		We showed the performance of these networks under the strain of reduced data samples, and in resolution reduction. The impact on performance as we iterate through the parameter space is relatively smaller for the eigenvalue features, but performance remains worse than the raw time-series.

		Finally, we provided a test-bed for further iteration on these sorts of prediction tasks, and opened up a discussion around sensor resolution, sample data size, downsampling, feature engineering, and CNNs. The comparison pipelines are easily extensible for further experimentation with this dataset or others. All of the code for feature engineering and testing is available on GitHub.
	
    	Variable resolution training has been employed on ImageNet (\cite{DAWNBench}) to dramatically reduce training time, it's interesting to consider the implications of explicitly controlling downsampling schemes for this ansatz. Larger scope, we have left open the question of ``multi-resolution'' sensor networks and the impact on geometric feature engineering and downsampling.

%% file: final_section_files/acknowledgements_and_impacts.tex
\section*{Acknowlegements}
The authors thank the anonymous reviewers from the NeurIPS 2020 TDA And Beyond Workshop.

We extend heartfelt appreciation to Will Chernoff for excellent feedback on the manuscript including a careful review of the possible implications of this work, leading to the broader impacts section. 

We also thank Dan Marthaler, Sven Schmit, and Hector Yee for feedback and comments on the manuscript. 

Janu Verma provided detailed feedback and recommendations on the exposition.

\section*{Broader Impacts}

As the primary application of study for these experiments lies within the medical space, both positive and negative applications jump to mind. Large, high-resolution datasets both for training and evaluation come at the benefit of those in developed and wealthy communities. Our research --  through its focus on developing methods robust to degradation -- provides an opportunity for an improvement in prediction methods in lower fidelity data regimes; i.e. methods designed with downsampling and data reduction in mind alleviate needs for larger and more complete datasets.

The present study examines EEGs and eye states, which could easily extend to other ailments. More accurate models for predicting seizure, for example, could greatly benefit those privileged enough to share characteristics with those used to train the model in the first place. But what of those not sampled?

As (\cite{Hall-rep}) has observed: Blacks, women, and the elderly have historically been excluded from clinical trial research. Such arrangements can lead to what (\cite{Veinot-intentions}) has referred to as intervention-generated inequalities (IGI), a social arrangement where one group gets better, while others don't. On top of their original ailments, groups left out are burdened with continued medical involvement and the associated costs (e.g. additional tests, transportation, childcare, and missed opportunities).

We offer the following suggestions for those in the medical industry hoping to combat some of this inequity: 1. Insist on multiple representative datasets including those from underrepresented groups -- incentivization where appropriate. 2. Identify and assist in eliminating barriers to involvement in data collection or diagnostics. 

%% file: final_section_files/appendices.tex
\section{Appendix: Persistent Homology}
\label{ph}

Fix $\epsilon \geq 0$, and let $X = \{ x_1, \ldots, x_n \}$ be an enumerated subset of $\mathbb{R}^m$. For $k = 0, 1, \ldots$ define $C_{\epsilon, k}$ to be the $\mathbb{R}$-vector space whose formal basis is given by all subsets of $X$ of the form 
\begin{equation*}
\{ (x_{i_0}, x_{i_1}, \ldots, x_{i_k}) \text{ such that } i_0 < i_1 < \cdots < i_k \text{ and } \norm{x_{i_{\alpha}} - x_{i_{\beta}}}  < \epsilon \text{ for all } \alpha, \beta = 0, 1, \ldots, k \}.
\end{equation*}
We take $C_{\epsilon, k}$ to be the zero vector space if there are no such subsets. There are linear maps $\partial_{k}: C_{\epsilon, k} \rightarrow C_{\epsilon, k-1}$ defined by

\begin{equation*}
    \partial_{\epsilon, k}(x_{i_0}, \dots, x_{i_k}) = \sum_{j = 0}^{k}(-1)^j(x_{i_0}, \ldots, \widehat{x_{i_j}}, \ldots, x_{i_k})
\end{equation*}

\noindent where $(x_{i_0}, \ldots, \widehat{x_{i_j}}, \ldots, x_{i_k})$ denotes the element $( x_{i_0},\ldots, x_{i_{j-1}}, x_{i_{j+1}},\dots, x_{i_k} ) \in C_{\epsilon, k-1}$. One can check that $\text{Im}(\partial_{\epsilon, k+1}) \subseteq \text{Ker}(\partial_{\epsilon, k})$. The \emph{$k^{th}\: \epsilon$-homology group} of $X$ is defined to be the vector space quotient $\text{H}_{\epsilon, k}(X) := \text{Ker}(\partial_{\epsilon, k})/\text{Im}(\partial_{\epsilon, k+1})$. The \emph{$k^{th}\: \epsilon$-Betti number} of $X$ is defined to be $\beta_{\epsilon, k}(X) := \text{dim}(\text{H}_{\epsilon, k}(X))$. Intuitively, $\beta_{\epsilon, k}(X)$ measures the number of $k$-dimensional holes in the point cloud $X$ at resolution $\epsilon$. We will typically write $\beta_k(\epsilon)$, and allow $\epsilon$ to vary over a fixed range $[0, r]$. We will also refer to $\beta_k(\epsilon)$ as an $\epsilon$-series. In the sequel, \emph{persistent homology} will in general refer to the collection of $\beta_k$'s, as they capture information about how the homology persists over varying resolutions. (c.f. \cite{Hatcher})

\section{Appendix: Downsampling}\label{downsampling}

\subsection{Time series downsampling}
    We consider a downsampling a selection of a subsequence of points, or a smaller set of points that summarize the timeseries. We assume the timeseries has $n+2$ points, and construct a downsample of $m+2$ points.
    
    \textbf{Naive Bucketing}: Select the first and last points of the timeseries; cover the the rest of the points with $m$ even-width intervals(up to integer rounding). We call this a \textit{bucketing}.  
    
    Consider a sequence of sequences:
    \begin{equation*}
        \left\lbrace\left\lbrace x_0 \right\rbrace, \left\lbrace x_1,\ldots,x_k\right\rbrace, \left\lbrace x_{k+1},\ldots,x_{2k}\right\rbrace, \ldots, \left\lbrace x_{(m-1)k+1},\ldots,x_{mk}\right\rbrace, \left\lbrace x_{n+1} \right\rbrace\right\rbrace
    \end{equation*}
    and for simplicity, call the sub-sequences $\left\lbrace{b_i}\right\rbrace_{0\leq i\leq n+1}$ such that $b_0 = \left\lbrace x_0 \right\rbrace$, $b_{n+1} = \left\lbrace x_{n+1} \right\rbrace$, and $b_j = \left\lbrace x_{(j-1)k + 1}\ldots x_{(j-1)k} \right\rbrace$, we refer to these as \emph{buckets}.
    
    \textbf{Dropout:} For each bucket, select the first point in the subsequence.
    
    \textbf{Bucket Averaging:} For each bucket in a naive bucketing, average the $x$ and $y$ coordinates and take this as the reprsentative point; take also the first and last points.

    For a bucket, we compute the 2-dimensional average of the points contained within: $\mu_j$. For convenience of notation, we write elements of $b_i$, as $\left\lbrace x^j_i\right\rbrace$. 
    
    \textbf{Largest-Triangle Three-Bucket Downsample (LTTB)} We compute the subsequence via the optimization problem:
    \begin{equation*}
        \textbf{compute } l_{i} = \argmax_{x^j_i} \mathbf{\triangle}\left({l_{i-1}, x^j_i, \mu_{i+1}}\right)\text{ such that } l_{0} = b_0 \text{ and } \mu_{n+1} = b_{n+1}.
    \end{equation*}
    The sequence $\left\lbrace l_0,\ldots, l_{n+1}\right\rbrace$ is the \emph{largest triangle three bucket downsample.} For more details and intuition around this construction we recommend the original paper.

    \begin{remark}
        This is computed via a recursive optimization process iterating through the buckets; a non-recursive formulation to find the global optima is also possible. The distinction between these two solutions is that in the recursive solution each optimization is conditioned on the previous bucket, where-as the global solution conditions on all buckets simultaneously.
    \end{remark}
    
\subsection{Dynamic downsampling}

In the above bucketing strategies, points in all regions of the time series are given equal weight in the downsample. Often times, the lagging-variance of a time series is not uniform across the time-domain. One might expect that regions of higher variance might warrant higher resolutions in the downsample, while low variance might require lower resolutions. A simple implementation of this idea, \textit{(inspired by \cite{sveinn})} is demonstrated in Algorithm \ref{alg:dynamic bucketing}(implementation included in Appendix). Downsampling methods are then applied to this bucketing of the timeseries.
    
\begin{algorithm}
  \caption{Variance weighted dynamic bucketing
    \label{alg:dynamic bucketing}}
  \begin{algorithmic}[1]
    \Require{$\mathcal{B}$ a naive bucketing, and $P$ an iteration count}
    \Statex
    \Function{dynamicbuckets}{$B:List[List[Float]$}
      \Let{$[b_j]$}{$\mathcal{B}$} 
      \For{$i \gets 1 \textrm{ to } P$}
        \For{$j \gets 1 \textrm{ to } m$}
            \Let{$S(b_j)$}{$SSE(OLS(b_j))$}
        \EndFor
        \Let{$z$}{$\argmax_j(S(b_j))$}
        \Let{$b^l_z$}{$\left\lbrace x^z_1, \ldots, x^z_{\floor{k/2}}\right\rbrace$}
        \Let{$b^r_z$}{$\left\lbrace x^z_{\floor{k/2}+1}, \ldots, x^z_k\right\rbrace$}
        \Let{$\mathcal{B}$}{$\mathcal{B}\setminus \left\lbrace b_z\right\rbrace \cup \left\lbrace b^l_z, b^r_z \right\rbrace$}
      \EndFor
      \For{$i \gets 1 \textrm{ to } P$}
        \For{$j \gets 1 \textrm{ to } m+P$}
            \Let{$S(b_j)$}{$SSE(OLS(b_j))$}
        \EndFor
        \Let{$a$}{$\argmin_a(S(b_a)+S(b_{a+1}))$}
        \Let{$b^*_a$}{$\left\lbrace x^a_1, \ldots, x^a_{k}\right\rbrace \cup \left\lbrace x^{a+1}_{1}, \ldots, x^{a+1}_k\right\rbrace$}
        \Let{$\mathcal{B}$}{$\mathcal{B}\setminus \left\lbrace b_a, b_{a+1} \right\rbrace \cup \left\lbrace b^*_a \right\rbrace$}
      \EndFor
      \State \Return{$\mathcal{B}$}
    \EndFunction
  \end{algorithmic}
\end{algorithm}

\begin{remark}
    Rather than serially splitting, and then combining buckets to arrive at the rebucketing, it's natural to ask how alternating these operations effects the result. The authors carried out several simulations of this technique and found that convergence to `stable' bucketing took place much more quickly, but produced far worse results with respect to total SSE.
\end{remark}

\subsection{Dynamic downsampling results}

We tested our dynamic downsampling strategies for the Laplacian eigenvalues features.

\begin{figure}[h]
\includegraphics[width=0.9\textwidth]{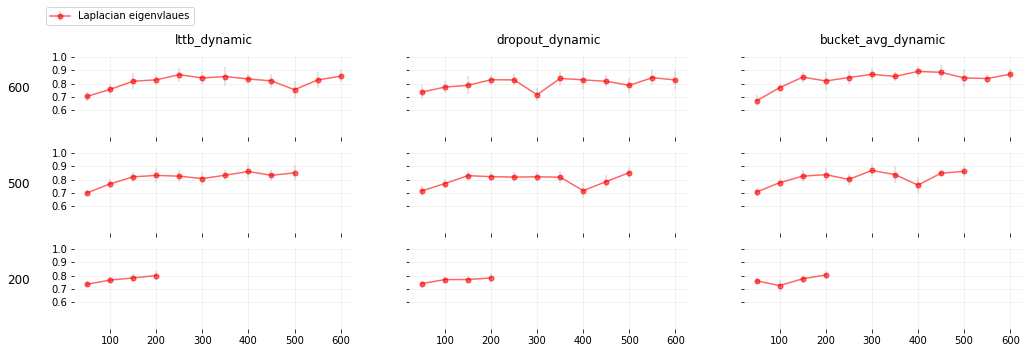}
\caption{Experiment results for dynamic bucketing downsampling methods. Diagram columns refer to downsampling method. Diagram rows correspond to initial length of time-series segment, with each point reflecting number of points after downsampling. Within rows, y-axis is accuracy on binary cross-entropy, x-axis is the number of points in the time-series samples.}
\label{fig:perf_dynamic}
\end{figure}

\section{Appendix: Cartoon version of this paper}

We thought it would be fun to capture the methods of this paper in a cartoon. The following image has a few numbered paths to walk through the three feature generation paths.

\begin{figure}[ht]
\includegraphics[width=0.9\textwidth]{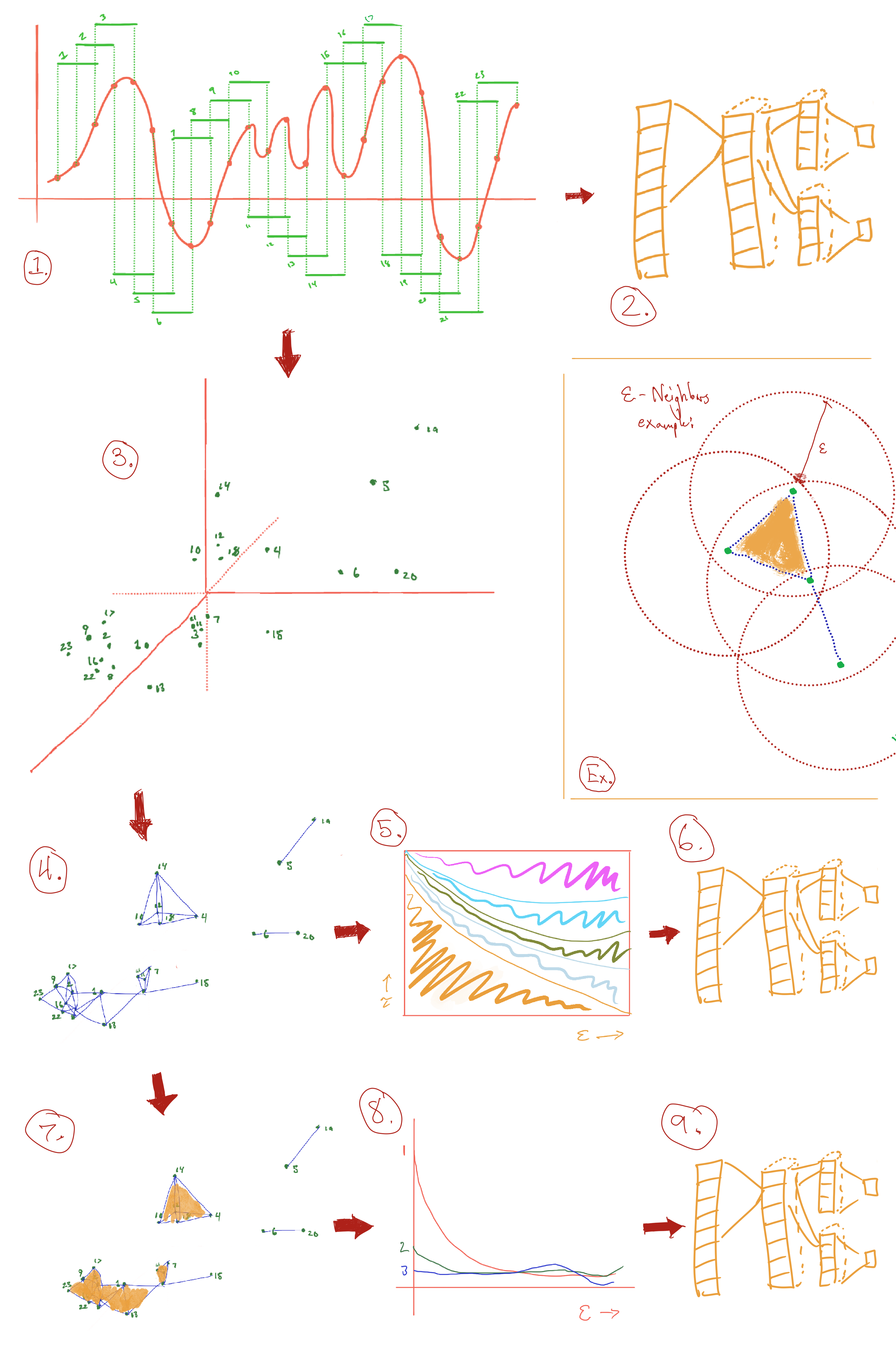}
  \caption{Three model pipelines. ($1\rightarrow 2$) Direct CNN on the raw features. ($1\rightarrow 3$) Takens' embedding. ($3\rightarrow 4$) generating the neighbor graph. ($4\rightarrow 5$) persistent Spectrum. ($4\rightarrow 7$) persistent Betti numbers. ($5\rightarrow 6$, $8\rightarrow 9$) CNN's from geometric features. (Ex.) $\epsilon$-neighbors for four points in turquoise.}
\label{fig:model_pipelines}
\end{figure}

%% file: neurips_2020.bbl
\begin{thebibliography}{19}
\providecommand{\natexlab}[1]{#1}
\providecommand{\url}[1]{\texttt{#1}}
\expandafter\ifx\csname urlstyle\endcsname\relax
  \providecommand{\doi}[1]{doi: #1}\else
  \providecommand{\doi}{doi: \begingroup \urlstyle{rm}\Url}\fi

\bibitem[Andrzejak et~al.(2001)Andrzejak, Lehnertz, Mormann, Rieke, David, and
  Elger]{eeg_data}
Ralph~G. Andrzejak, Klaus Lehnertz, Florian Mormann, Christoph Rieke, Peter
  David, and Christian~E. Elger.
\newblock Indications of nonlinear deterministic and finite-dimensional
  structures in time series of brain electrical activity: Dependence on
  recording region and brain state.
\newblock \emph{Phys. Rev. E}, 64:\penalty0 061907, Nov 2001.
\newblock \doi{10.1103/PhysRevE.64.061907}.
\newblock \url{https://link.aps.org/doi/10.1103/PhysRevE.64.061907}.

\bibitem[{Bronstein} et~al.(2017){Bronstein}, {Bruna}, {LeCun}, {Szlam}, and
  {Vandergheynst}]{geom_dl}
M.~M. {Bronstein}, J.~{Bruna}, Y.~{LeCun}, A.~{Szlam}, and P.~{Vandergheynst}.
\newblock Geometric deep learning: Going beyond euclidean data.
\newblock \emph{IEEE Signal Processing Magazine}, 34\penalty0 (4):\penalty0
  18--42, 2017.

\bibitem[Chazal and Michel(2017)]{Chazal2017AnIT}
Fr{\'e}d{\'e}ric Chazal and Bertrand Michel.
\newblock An introduction to topological data analysis: fundamental and
  practical aspects for data scientists.
\newblock \emph{ArXiv}, abs/1710.04019, 2017.

\bibitem[Chung and Graham(1997)]{chung1997spectral}
Fan~RK Chung and Fan~Chung Graham.
\newblock \emph{Spectral graph theory}.
\newblock American Mathematical Soc., 1997.

\bibitem[Covert et~al.(2019)Covert, Krishnan, Najm, Zhan, Shore, Hixson, and
  Po]{GCN-EEG}
Ian~C. Covert, Balu Krishnan, Imad Najm, Jiening Zhan, Matthew Shore, John
  Hixson, and Ming~Jack Po.
\newblock Temporal graph convolutional networks for automatic seizure
  detection.
\newblock In Finale Doshi-Velez, Jim Fackler, Ken Jung, David Kale, Rajesh
  Ranganath, Byron Wallace, and Jenna Wiens, editors, \emph{Proceedings of the
  4th Machine Learning for Healthcare Conference}, volume 106 of
  \emph{Proceedings of Machine Learning Research}, pages 160--180, Ann Arbor,
  Michigan, 09--10 Aug 2019. PMLR.
\newblock URL \url{http://proceedings.mlr.press/v106/covert19a.html}.

\bibitem[Dindin et~al.(2020)Dindin, Umeda, and Chazal]{TDAforArrhythmia}
Meryll Dindin, Yuhei Umeda, and Frederic Chazal.
\newblock Topological data analysis for arrhythmia detection through modular
  neural networks.
\newblock In Cyril Goutte and Xiaodan Zhu, editors, \emph{Advances in
  Artificial Intelligence}, pages 177--188, Cham, 2020. Springer International
  Publishing.
\newblock ISBN 978-3-030-47358-7.

\bibitem[Edelsbrunner et~al.(2000)Edelsbrunner, Letscher, and
  Zomorodian]{topo_persis}
H.~Edelsbrunner, D.~Letscher, and A.~Zomorodian.
\newblock Topological persistence and simplification.
\newblock In \emph{Proceedings of the 41st Annual Symposium on Foundations of
  Computer Science}, FOCS '00, page 454. IEEE Computer Society, 2000.
\newblock ISBN 0769508502.

\bibitem[Edelsbrunner et~al.(2002)Edelsbrunner, Letscher, and
  Zomorodian]{Edelsbrunner2000TopologicalPA}
Herbert Edelsbrunner, David Letscher, and Afra Zomorodian.
\newblock Topological persistence and simplification.
\newblock \emph{Discrete {\&} Computational Geometry}, 28\penalty0
  (4):\penalty0 511--533, Nov 2002.
\newblock ISSN 1432--0444.
\newblock \doi{10.1007/s00454--002--2885--2}.

\bibitem[Hall(1999)]{Hall-rep}
W.~D. Hall.
\newblock Representation of blacks, women, and the very elderly (aged> or= 80)
  in 28 major randomized clinical trials.
\newblock \emph{Ethnicity and disease}, 9(3):\penalty0 333--340, 1999.

\bibitem[Hatcher(2000)]{Hatcher}
Allen Hatcher.
\newblock \emph{{Algebraic topology}}.
\newblock Cambridge Univ. Press, Cambridge, 2000.
\newblock URL \url{https://cds.cern.ch/record/478079}.

\bibitem[Howard(2018)]{DAWNBench}
Jeremey Howard.
\newblock Now anyone can train imagenet in 18 minutes.
\newblock \url{https://www.fast.ai/2018/08/10/fastai-diu-imagenet/}, 2018.

\bibitem[Roy et~al.(2018)Roy, Ghosh, Bhattacharya, and Pal]{DBLP}
Prasun Roy, Subhankar Ghosh, Saumik Bhattacharya, and Umapada Pal.
\newblock Effects of degradations on deep neural network architectures.
\newblock \emph{CoRR}, abs/1807.10108, 2018.
\newblock \url{http://arxiv.org/abs/1807.10108}.

\bibitem[Scopigno et~al.(2004)Scopigno, Zorin, Carlsson, Zomorodian, Collins,
  and Guibas]{Scopigno04persistencebarcodes}
R.~Scopigno, D.~Zorin, Gunnar Carlsson, Afra Zomorodian, Anne Collins, and
  Leonidas Guibas.
\newblock Persistence barcodes for shapes, 2004.

\bibitem[Seo et~al.(2016)Seo, Defferrard, Vandergheynst, and Bresson]{gcrn}
Youngjoo Seo, Micha\"el Defferrard, Pierre Vandergheynst, and Xavier Bresson.
\newblock Structured sequence modeling with graph convolutional recurrent
  networks.
\newblock \emph{arXiv}, 2016.
\newblock URL \url{https://arxiv.org/abs/1612.07659}.

\bibitem[Steinarsson(2013)]{sveinn}
Sveinn Steinarsson.
\newblock Downsampling time series for visual representation.
\newblock Master's thesis, University of Iceland, 2013.

\bibitem[Umeda(2017)]{Umeda2017}
Yuhei Umeda.
\newblock Time series classification via topological data analysis.
\newblock \emph{Transactions of the Japanese Society for Artificial
  Intelligence}, 32\penalty0 (3):\penalty0 D--G72{\_}--12, 2017.
\newblock \doi{10.1527/tjsai.D--G72}.

\bibitem[Veinot~TC(2018)]{Veinot-intentions}
Ancker~JS. Veinot~TC, Mitchell~H.
\newblock Good intentions are not enough: how informatics interventions can
  worsen inequality.
\newblock \emph{J Am Med Inform Assoc.}, 25(8):\penalty0 1080--1088, 2018.

\bibitem[{Wang} et~al.(2017){Wang}, {Yan}, and {Oates}]{TSclass}
Z.~{Wang}, W.~{Yan}, and T.~{Oates}.
\newblock Time series classification from scratch with deep neural networks: A
  strong baseline.
\newblock In \emph{2017 International Joint Conference on Neural Networks
  (IJCNN)}, pages 1578--1585, 2017.

\bibitem[Zomorodian and Carlsson(2004)]{computingPH}
Afra Zomorodian and Gunnar Carlsson.
\newblock Computing persistent homology.
\newblock In \emph{Proceedings of the Twentieth Annual Symposium on
  Computational Geometry}, SCG '04, pages 347--356, New York, NY, USA, 2004.
  Association for Computing Machinery.
\newblock ISBN 1581138857.
\newblock \doi{10.1145/997817.997870}.

\end{thebibliography}
